\documentclass{conm-p-l}

\usepackage{amssymb, amsthm, amsmath,  amssymb, amsbsy,   amsfonts,  latexsym,  amsopn,   amstext, amsxtra,  euscript,   amscd}
\usepackage{amsmath, amssymb, amsthm}
\usepackage{booktabs}
\usepackage{tikz}
\usepackage{tikz,ifthen}
\usepackage{listings}
\usepackage{xcolor}

\usepackage{cite}
\usepackage[citation-order, msc-links]{amsrefs}

\usepackage{hyperref}
\hypersetup{
  colorlinks   = true,
  urlcolor     = blue,
  linkcolor    = blue,
  citecolor    = blue
}

\usepackage{algorithm}
\usepackage{algpseudocode}
\usepackage[all,poly]{xy}
\usepackage{tikz}
\usepackage{tikz,ifthen}
\usepackage{listings}

\newtheorem{prop}{Proposition} 
\newtheorem{lem}{Lemma} 
 
\newtheorem{defn}{Definition}

\usepackage[capitalise]{cleveref}

\crefname{thm}{Thm.}{}
\crefname{prop}{Prop.}{}
\crefname{lem}{Lem.}{}
\crefname{defn}{Def.}{}
\crefname{exa}{Example}{}
\crefname{exe}{Exercise}{}
\crefname{rem}{Rem.}{}
\crefname{cor}{Cor.}{}
\crefname{prob}{Problem}{}
\crefname{figure}{Fig.}{}
\crefname{table}{Table}{} 

\DeclareMathOperator\Jul{Jul}
\DeclareMathOperator\hyp{Hyp}

\DeclareMathOperator\re{Re }
\DeclareMathOperator\im{Im}

\DeclareMathOperator\wh{\mathfrak{h}}
\DeclareMathOperator\h{H}
\DeclareMathOperator\Sl{SL}

\newcommand\Z{\mathbb Z}
\newcommand\Q{\mathbb Q}
\newcommand\R{\mathbb R}
\newcommand\C{\mathbb C}

\newcommand\F{\mathcal F}            
\newcommand\J{\mathcal J}

\newcommand\Df{\mathfrak D_f}

\newcommand\D{\Delta}                
\newcommand\z{\xi}  	
\newcommand\zz{\zeta}    

\def\a{\alpha}
\def\b{\beta}

\def\O{\tilde{\mathcal O}}

\newcommand{\iu}{{i\mkern1mu}}

\def\V{\mathbb V}
\def\x{\mathbf{x}} 
      
\newcommand\y{\mathbf y}

\newcommand\w{\mathfrak w}

\newcommand{\norm}[1]{\left\lVert\mspace{1mu}#1\mspace{1mu}\right\rVert}
\newcommand{\abs}[1]{\left\vert\mspace{1mu}#1\mspace{1mu}\right\vert}

\DeclareMathOperator\wgcd{wgcd}   	

\begin{document}

\title{A Neurosymbolic Framework for Geometric Reduction of Binary Forms}

\author{Ilias Kotsireas}
\address{Department of Computer Science,  Wilfrid Laurier University, Waterloo, Ontario N2L 3C5, CANADA
  }
\email{ikotsireas@wlu.ca}

\author{Tony Shaska}
\address{Department of Mathematics and Statistics, Oakland University, Rochester, MI, 48309, USA}
\email{tanush@umich.edu}

\begin{abstract}
This paper compares Julia reduction and hyperbolic reduction with the aim of finding equivalent binary forms with minimal coefficients. We demonstrate that hyperbolic reduction generally outperforms Julia reduction, particularly in the cases of sextics and decimics, though neither method guarantees achieving the minimal form. We further propose an additional shift and scaling to approximate the minimal form more closely. Finally, we introduce a machine learning framework to identify optimal transformations that minimize the heights of binary forms. This study provides new insights into the geometry and algebra of binary forms and highlights the potential of AI in advancing symbolic computation and reduction techniques. The findings, supported by extensive computational experiments, lay the groundwork for hybrid approaches that integrate traditional reduction methods with data-driven techniques.
\end{abstract}


\maketitle

\section{Introduction} 
Reduction of integer binary forms is a classical problem in mathematics. There are many ways that the term \emph{reduction} is used. Here we will refer to it as the idea of picking a coordinate system such that the binary form has "small" coefficients. This is what is refer in \cite{2020-1} as \emph{Reduction A} versus \emph{Reduction B} which refers to picking the binary form with "smallest" invariants.

From now on, by \emph{reduction} of a binary form $f(x, y)$ defined over a field $k$, we will refer the process of picking a binary form $g(x, y)$ which is $k$-equivalent to $f(x, y)$ and has minimal coefficients. The only case that is fully understood and is part of the math folklore is the case of quadratic binary forms.

In his thesis \cite{julia} of $1917$, G. Julia introduced a reduction theory for binary forms with real coefficients (although explicit and complete answers were provided only in degrees three and four). To every binary form $f(x, y)$ with real coefficients, Julia associated a positive definite quadratic $\J_F$ which is called the \textit{Julia quadratic}. Cremona \cite{cremona} showed that the coefficients of $\J_F$ are polynomial values of the coefficients of $F$ and this does not happen for higher degree forms. Since positive definite quadratics parametrize $\mathbb{H}_2$, one obtains a well defined map $\zz$ from real binary forms to the upper half-plane. It is called \textit{the zero map} and it is $\Sl_2(\Z)$-equivariant. If $F$ is a real binary form, then $\zz (F)$ is a point in the hyperbolic convex hull of the roots of $F$ with non-negative imaginary part. A binary form is called \textit{reduced} if its image via the zero map is in the fundamental domain $\F$ of $\Sl_2(\Z)$.

In \cite{SC} Cremona and Stoll developed a reduction theory in a unified setting for binary forms with real or complex coefficients. A \textit{unique} positive definite Hermitian quadratic $\J_F$ is associated to every binary complex form $f(x, y)$. Since positive definite Hermitian forms parametrize the upper half-space $\mathbb{H}_3$, an extension of the zero map $\zz$ from binary complex forms to $\mathbb{H}_3$ is obtained. The upper half-plane $\mathbb{H}_2$ is contained in $\mathbb{H}_3$ as a vertical cross section (see the following section). When the form $f(x, y)$ has real coefficients, compatibility with complex conjugation forces $\zz(\J_F)\in \mathbb{H}_2$. It is in this sense that working in $\mathbb{H}_3$ unifies the theory of real and complex binary forms. A degree $n$ complex binary form $f(x, y)$ is called \textit{reduced} when its zero map value $\zz(\J_F)$ is in the fundamental domain of the action of the modular group $\Sl_2(\C)$ on $\mathbb{H}_3$.

In the works cited above, the term \textit{reduced binary form} means reduced in the $\Sl_2 (\Z)$ orbit. It is expected that the reduced forms have smallest size coefficients in such orbit. In \cite{2020-1}, the concept of height was defined for forms defined over any ring of integers $\O_K$, for any number field $K$, and the notion of \textit{minimal absolute height} was introduced and the author suggests an algorithm for determining the minimal absolute height for binary forms.

In \cite{2020-1} the authors introduce an alternative reduction method based purely on a geometric approach. For real cubics and quartics, Julia (\cite{julia}) uses geometric constructions to establish the barycentric coordinates $t_1, \dots , t_n$ of $\zz(F)$ in the hyperbolic convex hull of the roots of $F$. Geometric arguments are also used in \cite{SC} for the reduction of binary complex forms. In \cite{2020-1} reduction is based solely on a very special geometric point $\zz_{\mathcal C}(F)$ inside the hyperbolic convex hull of the roots of $F$, namely the \textit{hyperbolic centroid} of these roots. For a finite subset ${w_1,...w_n}\subset \mathbb{H}_2$, the \textit{hyperbolic centroid} is the unique point $x$ inside their hyperbolic convex hull which minimizes $\sum_{i=1}^n \cosh (d_H ({\mathbf x}, w_i)$ (here $d_H$ is the hyperbolic distance). To each real binary form $f(x, y)$ with no real roots, the alternative zero map associates the hyperbolic centroid of its roots. It is shown in \cite{2020-1} that this map is $SL_2 (\R)$ equivariant and different from Julia's, hence it defines a new reduction algorithm. Although zero maps are different, it seems that the effects of both reductions in decreasing the height are similar. Naturally, one would like to determine how different the zero maps are, or whether one can get examples where the reductions give different results.

The goal of this paper is to explore machine learning techniques, and more specifically neurosymbolic networks, to compare these two types of reduction and further investigate if any of them achieves the minimum height of the binary form. The simplest case would be that of binary sextics, and we will make use of machine learning methods used in \cite{2024-03} for such binary forms. While our methods and algorithms work for any degree, binary sextics and the database of \cite{2016-5, 2024-03} provide valuable examples where we can also see how the reduction affects the size of the invariants. 
We experiment with databases of irreducible quintics in \cite{2024-05} for the case when the binary form has one real root.

Our methods show that, in general, hyperbolic reduction is more effective than Julia reduction. However, it does not always achieve minimal height. In most cases, an \textbf{additional shift} is required to reach the minimal height through shifting. Since there is no known method to determine this additional shift, we employ machine learning techniques to further reduce the binary form and obtain its minimal height.

To conclude, the study of binary form reduction is not only a classical topic but also a rich intersection of geometry, algebra, and computational techniques. By applying modern machine learning frameworks as in \cite{sh-85, sh-86, sh-87, sh-88, sh-89, sh-90, sh-91,  sh-93, sh-94, sh-95, sh-97,  sh-99} we aim to provide new insights and algorithms that extend beyond traditional symbolic methods, paving the way for future advancements in this field.

\section{Preliminaries}
Let $k$ be a field and $k[x, y]$ the ring  of polynomials in two variables.    A  degree $n$ homogenous polynomial  $f\in k[x, y]$   can be written as 
\begin{equation}\label{eq1}
f(x, y)= \sum_{i=0}^n a_i x^i y^{d-i}
\end{equation}
for  $a_0, \ldots, a_n  \in k$.  Two homogenous polynomials $f$ and $g$  are called equivalent if $f(x, y)= \lambda \cdot g(x, y)$ for some $\lambda \in k^\star$.
Equivalence classes of homogenous polynomials are called \textbf{binary forms}.  The set of degree $n$ binary forms over $k$ will be denoted by $V_{n, k}$.  
There is a one to one correspondence between $V_{n, k}$ and the projective space $\mathbb{P}_k^n$.  Hence, sometimes we will denote the equivalence class of $f$ by $[a_0 : \ldots : a_n]$.  The \textbf{height} of $f$  (sometimes called the  \emph{naive height})    is defined as the height of $[a_0 : \ldots : a_n]\in \mathbb{P}_k^n$ and is denoted by $\h (f)$.  It is well-defined. 
When $k$ has characteristic zero and $f$ is primitive, then $\h (f) = \max \{\abs{a_i} \}$.

A \emph{quadratic form over $k$} is a function $Q: k^n \to k$ that has the form $Q(\mathbf{x})= \mathbf{x}^TA\mathbf{x}$,   where $A$ is a symmetric $n \times n$ matrix called the \emph{matrix of the quadratic form}.   Two quadratic forms $f(x, y)$ and $g(x, y)$ are called  \emph{$k$-equivalent} if one is obtained from the other by linear substitutions. 
In other words, $f(x, y) = f(ax+by, cx + dy)$,  for some $a, b, c, d \in k$.     Let $f$,  $g$ be quadratic forms and $A_f$,  $A_g$ their corresponding matrices, then  $f \sim g$ if and only if 
$A_f \sim A_g$.   

Let $k=\R$.     The binary quadratic form $Q$ is called \emph{positive definite} if $Q(\mathbf{x})>0$ for all nonzero vectors $\mathbf{x} \in \R^n$, and $Q$ is \emph{positive  semidefinite} if $Q(\mathbf{x})\geq 0$ for all  $\mathbf{x} \in \R^n$.    

Let $Q(x, y)= ax^2+ bxy + cy^2$. We will denote the  equivalence class of   $Q(x, y)$ by  $[a, b, c]$. The \emph{discriminant} of $Q$ is $\D= b^2-4ac$ and  $Q(x, y) $ is positive definite if $a>0$ and $\D <0$. 
Let 
\[V_{2, \R}^+= \left \{ Q(x, y) \in \R[x, y] \, \left | \frac{}{}\right. \, Q(x, y) \textit{ is positive definite  }\right \}.\]
Then   $\Sl_2(\R)$ acts    on $V_{2, \R}^+$  via 
 \[
\begin{split}
 \begin{bmatrix} 
 \a_1 \quad \a_2 \\ \a_3 \quad \a_4 
 \end{bmatrix} 
 \times 
  \begin{bmatrix} 
  x  \\ y
  \end{bmatrix}  
  &	\to Q(\a_1 x+\a_2 y,\a_3x+\a_4 y) =: Q^M (x, y)
\end{split}
\]
The discriminant $\D^\prime$ of $Q^M$ is   $\D^\prime=  (\det M)^2 \cdot \D = \D$. 
Hence,  $\D$ is fixed under the $\Sl_2(\R)$ action and the leading coefficient of the new form $Q^M$ will be $Q^M(1, 0)= Q(a, c) >0$.   Hence,  $ V_{2, \R}^+$ is preserved under this action.  
Consider the map  $\zz: V_{2, \R}^+ \to \mathbb{H}_2$
\begin{equation}\label{zero-map-real}
\begin{split}  
[a, b, c] & \mapsto \zz(Q)  = \frac{-b + \sqrt \D}{2a} 
\end{split}
\end{equation}
where $\re(\zz(Q)) =-\frac {b}{2a}$, and $\im (\zz(Q)) =\frac{ \sqrt{|\D|}}{2a}$.   
It is called the \textbf{zero map} (for quadratics)  and it is a bijection which  gives us   a one-to-one correspondence between positive definite quadratic forms and points in $\mathbb{H}_2$.

\begin{figure}[htbp]
\centering
\includegraphics[width=7cm, height=4cm]{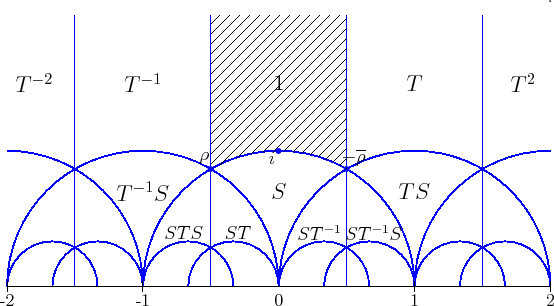}
\caption{The action of the modular group on $\mathbb{H}_2$.}
\label{fig:modular}
\end{figure}

The group  $\Gamma :=\Sl_2(\Z) /\{\pm I\}$is called the \textbf{modular group}.  It acts  on $V_{2, \R}^+$ as above. 
It also acts (from the right)   on $\mathbb{H}_2$ via  $\mathbb{H}_2    \times \Gamma  \to \mathbb{H}_2 $
\begin{equation}
\begin{split}
\left( z, M \right)   & \to  zM:=M^{-1} (z)    
\end{split}
\end{equation}
Note that the image is also in the upper half-plane, since
\[\im (M^{-1} (z ))= \det(M^{-1}) \cdot  \frac{\im (z )}{  \norm{\a_1-a_3z}^2}.\]
This action has a  fundamental domain  $\F$ 
\[\F= \left\{\frac{}{} z \in \mathcal H_2  \left| \frac{}{} \right.    \, |z|^2 \geq 1  \, \text{ and } \, |Re(z)| \leq 1/2 \right\}\]
as   displayed in  \cref{fig:modular}.

%
%
The zero map   is  $\Gamma$-equivariant, i.e.  $\zz(Q^M)=M^{-1}\zz(Q)$ 
\[
\xymatrix{
    V_{2, \mathbb{R}}^+ \ar[r]^{\zz} \ar[d]_{M} & \mathbb{H}_2 \ar[d]^{M^{-1}} \\
    V_{2, \mathbb{R}}^+ \ar[r]^{\zz} & \mathbb{H}_2
}
\]
%

%
We   define the quadratic $Q=[a, b, c]$ to be a \textbf{reduced quadratic } if $\zz(Q) \in \F$.

\begin{lem}\label{red-quad}  The following are true:
\begin{enumerate}
\item      $Q \in V_{2, \R}^+$ is  reduced if and only if  $|b| \leq a \leq c$. 
\item    Let $Q$ be a reduced form with   $\D=-D$. Then  $b \leq \sqrt{ D/ 3}$.
\item The number of reduced forms with    $\D=-D$  is finite.
\item Every $Q \in V_{2, \R}^+$   is equivalent to a reduced one. 
\end{enumerate}
\end{lem}
  
Two reduced binary quadratics are equivalent only in the following two cases  $[a, b, a] \sim [a, -b, a]$ or $[a, a, c] \sim [a, -a, c]$.    Let $\D < 0$ be fixed. Then the  class number $h(\D)$ is equal to the number of primitive reduced forms of discriminant $\D$. 
 

\subsection{The hyperbolic plane $\mathbb{H}_2$} 
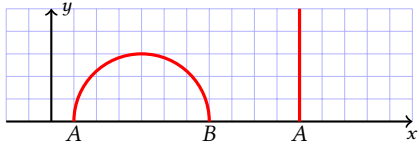
\begin{figure}[h!] 
   \centering
\begin{tikzpicture}[help lines/.style={blue!30,very thin},scale=0.3]
    \draw [help lines] (-2, 0) grid (16, 5);
    \draw[->, thick] (-2, 0) -- (16, 0) node[below]{\footnotesize $x$};
    \draw[->, thick] (0, 0) -- (0, 5) node[right]{\footnotesize $y$};
\draw[red, very thick ] (7, 0) arc (0:180:3cm);
    \draw[-, color=red,very thick] (11, 0) -- (11, 5);
    \node at (1, -0.5) {$A$};
        \node at (7, -0.5) {$B$};
            \node at (11, -0.5) {$A$};
\end{tikzpicture}
   \caption{Ideal points and the corresponding geodesics}
   \label{fig1}
\end{figure}

The upper-half plane equipped with the Riemanian metric 
\[
ds^2=\frac{dx^2+dy^2}{y^2}
\]
 is one of the models of the two dimensional hyperbolic space. 
The geodesics of the Riemaniann manifold $\mathbb{H}_2$, i.e the hyperbolic equivalents of Euclidean straight lines, are either semicircles $C_{a,b}$ with diameter from $A(a,0)$ to $B(b,0)$ on the real axis, or the vertical rays $C_a$ with origin at $x=a$. In the standard literature, the points $A(a,0), B(b,0)$ are called {\bf the ideal points} of the geodesic $C_{a,b}$, likewise $A(a,0)$ and $\infty$ are the ideal points of $C_a$. The ideal points of the geodesic live in the boundary of $\mathbb{H}_2$;  see \cref{fig1}.

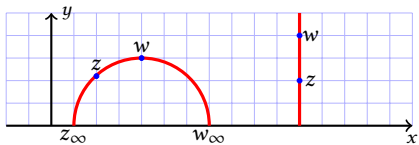
\begin{figure}[!h] 
   \centering
\begin{tikzpicture}[help lines/.style={blue!30,very thin},scale=0.3]
    \draw [help lines] (-2, 0) grid (16, 5);
    \draw[->, thick] (-2, 0) -- (16, 0) node[below]{\footnotesize $x$};
    \draw[->, thick] (0, 0) -- (0, 5) node[right]{\footnotesize $y$};
\draw[red, very thick ] (7, 0) arc (0:180:3cm);
    \draw[-, color=red,very thick] (11, 0) -- (11, 5);
    \node at (11.5, 2) {$z$};
    \node at (11.5, 4) {$w$};
    \foreach \x/\y in {   4/3, 2/2.2, 11/2, 11/4}
    \filldraw[blue] (\x, \y) circle(3pt);
    \node at (2, 2.7) {$z$};
    \node at (4, 3.5) {$w$};
    \node at (1, -0.5) {$z_{\infty}$};
    \node at (7, -0.5) {$w_{\infty}$};
\end{tikzpicture}
   \caption{Hyperbolic distance between two points $z$ and $w$}    
   \label{fig2}
\end{figure}

Let $z=x+{\bf i}y$, $w=u+{\bf i}v$ and  $z_{\infty}, w_{\infty}$ be the ideal points of the geodesic through $z,w$, where $z_{\infty}$ is the one closer to $z$; see  \cref{fig2}.  

The hyperbolic distance is defined as
%
\[ 
d_H(z,w)=
\ln\left(\frac{|z-w_{\infty}|}{|w-w_{\infty}|}\frac{|w-z_{\infty}|}{|z-z_{\infty}|}\right).
\]
For $x=u$ and $y<v$, the geodesic is the vertical ray $C_x$. In this case  $z_{\infty}=(x,0)$, $w_{\infty}=\infty$ and 
\[ d_H(z,w)=\ln \left(\frac{v}{y}\right).
\] 
For $A(a,0)$ and $z=x+{\bf i}y\in \mathbb{H}_2$, define 
\[ d_H(A,z):=\ln\left(\frac{(x-a)^2+y^2}{y}\right).\]
%

\begin{prop} 
Let $A$ be one of the ideal points of a geodesic that passes through $z=x+y{\bf i},w=u+v{\bf i}\in \mathbb{H}_2$. Then 
\[d_H(z,w)=|d_H(A,z)-d_H(A,w)|.\] 
\end{prop}

\subsection{$\mathbb{H}_2$ as a parameter space for positive definite real quadratic forms.} 
 $\mathbb{H}_2$ parametrizes binary quadratic forms with discriminant $\Delta<0$ and $a>0$, while its boundary parametrizes those with discriminant $\Delta=0$.  In \cite{e-sh} it was proved that: 
  
\begin{prop}
Let $\overline{\mathbb{H}}_2=\mathbb{H}_2\cup \partial \mathbb{H}_2=\mathbb{H}_2\cup\R\mathbb{P}^1$ and $\omega_1,\omega_2\in \overline{\mathbb{H}}_2$. The quadratics of the form $$s Q_{\omega_1}+t Q_{\omega_2},s>0,t>0, s+t=1$$ parametrize the hyperbolic segment that joins $\omega_1$ and $\omega_2$. 
\end{prop}

This proposition is generalized by induction as follows  (see \cite{e-sh}):

\begin{prop}\label{convex hull}
Let $\omega_1,\omega_2,...,\omega_n\in \overline{\mathbb{H}}_2$ such that for all $i$, $\omega_i$ is not in the hyperbolic convex hull of $\omega_1,\omega_2,...,\omega_{i-1}$. Then the convex hull of  $\omega_1,\omega_2, \dots ,\omega_n$ parametrizes the linear combinations $\sum_{i=1}^n\lambda_iQ_{\omega_i}$ with $\lambda_i\geq 0$ and $\sum_{i=1}^n\lambda_i=1$.
\end{prop}

\subsection{The hyperbolic three dimensional space $\mathbb{H}_3$} 
As a set, $\mathbb{H}_3=\C \times \R^+$. Points of $\mathbb{H}_3$ will be written in the form $z+t{\bf j}$ where $z\in \C$ and $t>0$.  The equation $t=0$ represents the floor $\C$ of $\mathbb{H}_3$. The hyperbolic space $\mathbb{H}_3$  is foliated via horospheres %
\[H_t:=\{z+t{\bf j}:~z\in \C\}\] 
which are centered at $\infty$ and indexed by the height $t$ above $\partial \mathbb{H}_3=\mathbb{P}^1\C$. The algebra of $\mathbb{H}_3$ is not commutative. Then
${\bf j}^2=-1$, ${\bf i}{\bf j}=-{\bf j}{\bf i}$,  ${\bf j}z=\bar{z}{\bf j}$.
The notion of complex modulus extends to $\mathbb{H}_3$:  $|z+t{\bf j}|=|z|^2+t^2$. There is a natural isometrical inclusion map $\mathbb{H}_2\rightarrow \mathbb{H}_3$ via $x+{\bf i}t\rightarrow x+{\bf j}t$. Invariant elements of $\mathbb{H}_3$ under the partial conjugation 
\[z+{\bf j}t \mapsto \bar z+{\bf j}t\]
are  elements of $\mathbb{H}_2$. The hyperbolic metric is 
\[ds^2=\frac{|dz|^2+dt^2}{t^2}.\] 
The geodesics are either semicircles centered on $\C$ and perpendicular to $\C$, or rays $\{z_0+{\bf j}t\}$ perpendicular to $\C$. 

\begin{figure}[h!] 
   \centering
   \includegraphics[width=3in]{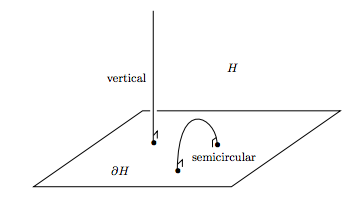} 
   \caption{Geodesics in $\mathbb{H}_3$}
   \label{fig5}
\end{figure}

For $\omega=z+t{\bf j}\in \mathbb{H}_3$ and $w+0{\bf j}\in \C$ in the floor, define 
\[
d_H(\omega,w):=\frac{|z-w|^2+t^2}{y}.
\]

\begin{prop}
If one of the ideal points of the geodesic through $\omega_1,\omega_2$ is at $w$, then $$d_H(\omega_1,\omega_2)=|d_H(\omega_1,w)-d_H(\omega_2,w)|.$$ 
\end{prop}

There is a right action of $\Sl_2(\C)$ on $\mathbb{H}_3$. If $M\in \Sl_2(\C)$ and $M^{-1}= \begin{pmatrix}  a & b \\ c & d  \end{pmatrix}$, its action is described as follows 
\[(z+{\bf j}t)\cdot M=M^{-1}(z+{\bf j}t)=[a(z+{\bf j}t)+b][(c(z+{\bf j}t)+d]^{-1},\] 
where the inverse indicates the right inverse in the non commutative structure of $\mathbb{H}_3$.  Note that for $t=0$ we get the standard $\Sl_2(\C)$-action on the boundary $\C\mathbb{P}^1$ of $\mathbb{H}_3$.

\begin{lem} The action of $\Sl_2(\C)$ on $\mathbb{H}_3$ can be written in the form
\[
(z+{\bf j}t)\cdot M=\frac{(az+b)\overline{(cz+d)}+a\bar ct^2+{\bf j}t}{|cz+d|^2+|c|^2t^2}.
\]
\end{lem}

\subsubsection{ $\mathbb{H}_3$ as a parameter space for positive definite Hermitian quadratic forms}
 Let $$H(x,y)=a|x|^2-bx\Bar y-\bar b\bar xy+c|y|^2, a,c\in \R$$ be a Hermitian quadratic form with homogeneous variables $[x,y] \in \mathbb{P}^1\C$. Notice that the values of $H(x,y)$ are always real. Let $\Delta=ac-|b|^2$ be its discriminant. Then $$H(x,y)=a[x-(\bar b/a)y]^2+(\Delta/a)y^2,$$ hence $H(x,y)>0$ for all $(x,y)$ when $\Delta >0,a>0$. Such a form is called {\bf positive definite.}  Denote the set of all positive definite Hermitian forms by $V^+_{2,\C}$. There is an $\Sl_2(\C)$ action on $V^+_{2,\C}$ similar to the real case. The natural $\Sl_2(\R)$ equivariant inclusion $\psi: V^+_{2,\R}\rightarrow V^+_{2,\C}$ via $$\psi(ax^2-2bxy+cy^2)=a|x|^2-bx\bar y-\bar b\bar xy+c|y|^2,$$ gives rise to an extension of the zero map.

\begin{defn}
The zero map $\zz: V^+_{2,\C}\rightarrow \mathbb{H}_3$ is defined via
\begin{equation} \label{eq: hermitian zero map}
\zz(a|x|^2-bx\bar y-\bar b\bar xy+c|y|^2)=\frac{\bar b}{a}+{\bf j}\frac{\sqrt \Delta}{a}
\end{equation}
\end{defn}

\begin{prop} \label{complex invariant}
The map $\zz$ is $\Sl_2(\C)$ equivariant.
\end{prop}
 
The hyperbolic space $\mathbb{H}_3$ parametrizes (up to a constant factor) positive definite ($\Delta>0,a>0$) Hermitian forms via the inverse map $$\zz^{-1}(\omega)=\zz^{-1}(z+{\bf j}t)= |x|^2-\bar{z}\bar xy-zx\bar y+(|z|^2+t^2)|y|^2=H_{\omega},$$ while the boundary $\C\mathbb{P}^1=\C\cup{\infty}$ of $\mathbb{H}_3$ parametrizes the decomposable ($\Delta=0$) Hermitian forms \[H_{\beta}=(x-\bar \beta y)(\bar x-\beta \bar y)~\text{for}~\beta\in \C,~\text{and}~H_{\infty}=|y|^2,\] Just as in the case of $\mathbb{H}_2$, we have the following proposition:

\begin{prop} \label{convex hull H3}
Let $\overline{\mathbb{H}}_3=\mathbb{H}_3\cup \partial \mathbb{H}_3=\mathbb{H}_3\cup \C\mathbb{P}^1$. The hyperbolic convex hull of $\omega_1,\omega_2,...,\omega_n\in \overline{\mathbb{H}}_3$ parametrizes Hermitian forms $\sum_{i=1}^n \lambda_i H_{\omega_i}$ with $\lambda_i>0$ for $i=1,2,...,n$ and $\sum_{i=1}^n \lambda_i=1$. 
\end{prop}

The equivariant connection between the geometry of hyperbolic spaces and the algebra of positive definite forms, which extends to the boundary as well, can be expressed in the following equivariant commutative diagram:
\[
\xymatrix{
 V^+_{2,\R}   \ar@{->}[d]  \ar@{->}[r]^\zz      & \mathbb{H}_2  \ar@{->}[d]^{}  \\
 V^+_{2,\C}        \ar@{->}[r]^\zz                   & \mathbb{H}_3 \\
}
\]
Next, we will see how to use the equivariance of the zero map to construct a reduction method.

\section{Reduction of binary forms}

\subsection{Julia  reduction} 
Let $f(x,y) \in \R[x,y]$ be a degree $n$ binary form  given as in \cref{eq1}.
 The form is factored as 
\begin{equation}\label{deg-n-real-factored}
f(x,1) = \prod_{i=1}^r (x-\a_i) \cdot\prod_{i=1}^s (x-\b_i)(x-\bar \b_i).
\end{equation}
where    $\a_1, \ldots , a_r \in \R $  and  $\b_1, \ldots , \b_s$ are   in the upper half complex plane, denoted by $\mathbb{H}_2$.  
The ordered pair $(r, s)$   is called the \emph{signature} of  $f$. 
We associate to $f$    

the quadratic  $Q_f$  
\begin{equation}\label{Q_f}
Q_f (x, y) =   \sum_{i=1}^r t_i^2(x-\a_i y)^2 +    \sum_{j=1}^s 2u_j^2 (x-\b_j y)(x-\bar \b_j y),
\end{equation} 
where $t_i$, $u_j$ are   to be determined.    
Let $\b_j=a_j + b_j  \iu$, for $j=1, \dots , s$. 
The discriminant of $Q_f$ is a degree 4 homogenous polynomial in $t_1, \dots t_r,$ $u_1, \dots , u_s$.  We would like to pick values for $t_1, \dots t_r, u_1, \dots , u_s$ such that this discriminant is square free and minimal.  

$Q_f$ is a positive definite quadratic form with discriminant $\Df$;   which is expressed in terms of the root differences; see \cite{2020-1}.  Hence, $\Df$ is fixed by all the transpositions of the roots. 
Indeed $\Df^n$ is an invariant of the binary form $f$.     
  We define the $\theta_0$ of $f$ as 
\begin{equation}\label{Julia-def}
\theta_0 (f) = \frac{a_0^2\cdot |\Df|^{n/2}}{\prod_{i=1}^rt_i^2\, \prod_{j=1}^s u_j^4  }.
\end{equation}

Consider $\theta_0 (t_1, \dots, t_r, u_1, \dots, u_s)$  as a multivariable function in the variables $t_1, \dots , t_r, u_1, \dots , u_s$.  We would like to pick these variables such   $\Df$ is minimal.  This is equivalent to $\theta_0(t_1, \dots, t_r, u_1, \dots, u_s)$ obtaining a minimal value.   

\begin{prop}\label{unique-min}
The function  $\theta_0 : \R^{r+s} \to \R$  obtains a minimum at a unique point $(\bar t_1, \dots, \bar t_r,  \bar u_1, \dots, \bar u_s)$. 
\end{prop}

Choosing $(\bar t_1, \dots, \bar t_r,  \bar u_1, \dots, \bar u_s)$  that   make $\theta_0$ minimal  gives a unique positive definite quadratic $Q_f (x, y)$.  We call this unique quadratic $Q_f (x, y)$ for such a choice of  $(\bar t_1, \dots, \bar t_r,  \bar u_1, \dots, \bar u_s)$  the \textbf{Julia quadratic} of $f(x, y)$ and   denote it by $\J_f (x, y)$.   
The quantity  
\[
\theta_f:= \theta_0(\bar t_1, \dots, \bar t_r,  \bar u_1, \dots, \bar u_s)
\]
 is called the \textbf{Julia invariant}.   

\begin{lem}\label{theta-invariant}
Let $f\in V_{n, \R}$. Then    $\theta_f$ is an   $\Sl_2 (\R)$- invariant  and   $\J_f$ is an $\Sl_2 (\R)$  covariant  of order 2. 
\end{lem} 
 
Hence we have   $\V_{n, R} \to V_{2, \R}^+ \to \mathbb{H}_2$ via 
\[
f  (x, y) \to \J_f (x, y) \to \zz (\J_f  )  =: \zz (f)
\]
The map 
\[
\zz : V_{n, \R} \to \mathbb{H}_2
\]
 is  called  \textbf{Julia zero map} and it is  $\Sl_2 (\R)$-equivariant; see \cite{julia, SC}.
The zero map extends to 
\[
\z: V_{n, \C}    \rightarrow \mathbb{H}_3
\]
 via $\z (f)=\z (\J_f)\in \mathbb{H}_3$, a point in the hyperbolic convex hull of the roots of $f$. 
The form $f$ is called \textbf{ Julia reduced} if $\z(f)$ is in the fundamental domain $\mathcal F$ of $\Sl_2(\C)$.

If Julia quadratic preserves the height, 
then Julia reduction  would give a form with minimal height. However, this 
is not true as shown in \cite{2020-1} for cubics and it will shown again in the coming section.


\subsection{Hyperbolic reduction} \label{hyb-sec}

Hyperbolic reduction was introduced in \cite{2020-1} when authors showed that  using the hyperbolic centroid  for the zero map instead of the center of mass gives different results from 
  Julia reduction.  Below, we briefly describe this reduction in detail, since it is less known than Julia reduction and provide explicit formulas how to compute the hyperbolic centroid for a set of points in the upper complex plane.
  For further details see \cite{2020-1}.
  
\begin{defn} The hyperbolic centroid, or simply centroid, 
\[
{\mathcal C_{H}}(\alpha_1,\alpha_2,...,\alpha_n)
\]
 of the collection $\{\alpha_j\in \mathbb{H}_2~ | j=1,2,...,n\}$ is the unique point $t+{\bf i}u\in \mathbb{H}_2$ that minimizes $$\sum_{j=1}^n \frac{(t-x_j)^2+(u-y_j)^2}{uy_j}.$$ 
\end{defn}

\begin{prop} \label{centroideqns} 
The centroid  ${\mathcal C_{H}}=t+{\bf i}u\in \mathbb{H}_2$ of $\alpha_1,\alpha_2,...,\alpha_n$ satisfies
\begin{equation}\label{centroid}
\begin{split}
 \displaystyle{t} & \displaystyle{=\sum_{i=1}^n\left(\frac{ y_1 y_2 \cdots y_{i-1} y_{i+1} \cdots y_n}{\mathfrak s_{n-1}(y_1,y_2,...,y_n)}\right)x_i}  \\
\displaystyle{|{\mathcal C_{H}}|^2}  & =  \displaystyle{ \sum_{i=1}^n \left(\frac { y_1 y_2 \cdots y_{i-1} y_{i+1} \cdots y_n }   {\mathfrak s_{n-1}(y_1,y_2,...,y_n) }\right)|\alpha_i|^2}  \\
\displaystyle{Q_{{\mathcal C_{H}}}(x, y)} & = \displaystyle{\sum_{i=1}^n \left(\frac { y_1 y_2 \cdots y_{i-1} y_{i+1} \cdots y_n } {\mathfrak s_{n-1}(y_1,y_2,...,y_n)}\right)Q_{\alpha_i}(x, y)}. \\
\end{split}
\end{equation}
\end{prop}
 It follows that as a point in the hyperbolic convex hull of $\alpha_1,\alpha_2,...,\alpha_n$, the centroid is represented by the linear combination positive definite quadratic \[Q_{{\mathcal C}_{H}}(x, y)=\displaystyle{ \sum_{i=1}^n \left(\frac { y_1 y_2 \cdots y_{i-1} y_{i+1} \cdots y_n }   { \sigma_{n-1}(y_1,y_2,...,y_n) }\right)Q_{\alpha_i}(x, y)}.\]

  All equations in \cref{centroid} are described in terms of the function  defined by
\[
\begin{split}
 \psi : \R^n \times \R_{>0}^n  &  \mapsto \R \\
\psi\left(     \left( x_1, \dots , x_n \right), \left( y_1, \dots , y_n \right) \right)  &=\sum_{i=1}^n \left(\frac { y_1 y_2 \cdots y_{i-1} y_{i+1} \cdots y_n }   { \mathfrak s_{n-1}(y_1,y_2,...,y_n) } \right)x_i. \\
\end{split}
\]
The function $\psi$ has symmetries and is a convex linear combination of $x_i$'s with weights that depend only on $y_1,...,y_n$. It is probably a well-known and standard function in areas where symmetries and group actions are relevant.


Let $V^+_{2n,\R}(0,n)$ denote binary forms of degree $2n$ with real coefficients and no real roots. Every $f(x, y)\in V^+_{2n,\R}(0,n)$ can be factored 
\[
f(x, y)=\prod_{j=1}^n Q_{\alpha_j}(x, y)
\] 
where 
$ \alpha_j=x_j+{\bf i}y_j$,     and 
\[
Q_{\alpha_j}(x, y)=(X-\alpha_jZ)(X-\overline{\alpha_j}Z). 
\]
The \textbf{centroid zero map} 
\[
\z_{\mathcal C}: V^+_{2n,\R}(0,n)\rightarrow \mathbb{H}_2
\]
 is defined via 
\[
\z_{\mathcal C}(F):=\mathcal C_{H} ={\mathcal C}_{H}      (\alpha_1,\alpha_2,...,\alpha_n).
\] 
The form
\[
\begin{split}
\J^{\mathcal C}_f  &  :=       (x-{\mathcal C_{H}} \, y)  (x-\overline{\mathcal C_{H}} \, y)   \\
& =  \sum_{j=1}^n \left(\frac { y_1 y_2 \cdots y_{j-1} y_{j+1} \cdots y_n } {\mathfrak s_{n-1}(y_1,y_2,...,y_n)}\right)Q_{\alpha_j}(x, y)
\end{split}
\] 
is called the \textbf{centroid quadratic} of $F$.

The reduction theory based on the centroid proceeds as in the case of Julia reduction.
 Let $f(x, y)$ be a real binary form with no real roots. If $\z_{\mathcal C}(f)\in \F$ then $f$ is reduced. Otherwise, let $M\in \Sl_2(\R)$ such that $M^{-1}\z_{\mathcal C}(f) \in \F$. The form $f$ reduces to $f^M(x, y)$.

In \cite{e-sh} it was given a formula for computing the hyperbolic centroid:

\begin{prop}\label{hyb_centroid}
Let $F(X, Z)$ be a totally complex form factored over $\R$ as below
\[ F(X, Z)= \prod_{i=1}^n (X^2 + a_i X Z  + b_i Z^2) \]
Denote by $d_i = \sqrt{4b_i-a_i^2}$, for $i=1, \dots ,  n$ the discriminants for each factor of $f(x, y)$. Let 
\[
\mathfrak s_{n-1}= \sum_{i=1 }^r d_1 \cdots  d_{i-1}\hat{d_i}d_{i+1}\cdots d_r,
\]
 where $\hat{x}$ denote a missing $x$, and 
\[  {\bf a}=(a_1,...,a_n),   \; \; {\bf b}=(b_1,...,b_n), \; \; {\bf d}=(d_1,...,d_n).\]
The centroid quadratic of $f(x, y)$ is given by 
\[\displaystyle{\J^{\mathcal C}_F=\sum_{i=1}^n \left(\frac { d_1 d_2 \cdots d_{i-1} d_{i+1} \cdots d_n } {\mathfrak s_{n-1}}\right)(X^2+a_iXZ+b_iZ^2)}.\] 
The centroid zero map $\z_{\mathcal C}(F)=t+{\bf i}u\in \mathbb{H}_2$ is given by
\begin{small}
\[
\begin{split}
&  t  =  - \frac 1 {2}  \,     \sum_{i=1 }^n \frac{d_1 \cdots  d_{i-1}d_{i+1}\cdots d_n}{\mathfrak s_{n-1}}a_i =\psi({\bf d},{\bf a}) \, ,~ \\
&   u^2   = \frac  1   { 4 \mathfrak s_{n-1}^2}     \, \prod_{i=1}^n d_i      \,  
   \left( \mathfrak s_{n-1} \, \sum_{i=1}^n d_i  +      \sum_{i}^n \, d_1 \cdots \hat{d_i}\cdots \hat{d_j}\cdots d_n  \left( a_i - a_j  \right)^2    \right)    \\
& |\z_{\mathcal C}(F)|^2  =     \sum_{i=1 }^n \frac{d_1 \cdots  d_{i-1}d_{i+1}\cdots d_n}{\mathfrak s_{n-1}}b_i=\psi({\bf d},{\bf b})  \, .
%
\end{split}
\]
\end{small}
The reduction is defined over $\Q(d_1, d_2, ...,d_n)$
\end{prop}

Even for hyperbolic reduction, similarly to the Julia reduction, the main question is the same:  does the reduced binary form have minimal height?  An affirmative answer to this question is very unlikely (as it was the case for Julia reduction).  Moreover, we would like to know which one performs better in general.  This will be investigated next. 

\section{A database of binary forms}\label{data}
Next we want to construct a database of binary forms so we can possibly discover properties of our reduction methods and design the best possible model for reduction. 
In building a database of binary forms we can follow two main methods.

First, we can use databases of binary forms  from \cite{2024-03}.   Such  degree $n> 2$    binary forms are  points in the projective space $\mathbb{P}^n$.  However, because of the way such databases were constructed most of those binary forms  have minimal hight and would be useless to us for illustrating reduction methods.  
In order to have this data  $\{ f\}_S$, for some index set $S$,   useful for  training, we can randomly act on each binary form   with random matrices $M \in \Sl_2 (\Z)$.  The new data $\{ f^M \}_S$,
will not, in general,  have binary forms with minimum heights.   However, we can design a machine learning model based on $\{ f^M \}_S$, and do the training of this model based on $\{ f\}_S$.

Second , we start with roots in the hyperbolic plane $\mathbb{H}_2$.  We create a database of   $n$-gons with vertices $\a_1, \ldots , \a_r \in \mathbb{H}_2$.  For simplicity of the argument here we assume we have no real roots, even though the method can be easily extended in this case. 
Thus our binary form will be   
\[
f(x, y) = \prod_{j=1}^r \left(x^2- 2 \re(a_j)  x y + ||{\a_j} ||^2 y^2 \right).
\]
Binary forms of this type are called \emph{totally complex forms}.
In order to have $f(x, y)$ with integer coefficients we can further assume that $\a_1, \ldots , \a_r \in \mathbb{H}_2$ are Gaussian integers.   

To control the location of the polygons we can assume that the roots $\a_1, \ldots , \a_n$ are always picked between radii $r_1$ and $r_2$.  This assures that 
we don't take ones close to the fundamental domain (so the affects of the reduction are more visible)  or we don't have floating issues in the case of very large coordinates. 
The main question here becomes how to pick $r_1$ and $r_2$ so we can get a database of preferable size. 

The number of Gaussian integers in this region is roughly $\pi (r_2^2 - r_1^2)$  is related to the famous \emph{Gauss circle problem}.  Hence, we can always have some estimate of how many points we will have in the region and therefore the number of $n$-gons, which is much bigger then the number of points in the region. 
As you will see below, there are $37\, 090\, 735$ triangles for $r_1-1$ and $r_2=20$ and 
$8 \, 936 \, 928$   pentagons for $r_1=1$ and $r_2=7$.

These two very different approaches of creating a database of binary forms are mostly forced upon us by the strategies of building a training model.  

The algebraic approach would be to ignore the geometry (roots of binary forms) and express the Julia invariant   in terms of the coefficients of the form $a_0, \ldots , a_n$.  Since the Julia invariant is an invariant of the form then it must be expressed in terms of such coefficients.  We can design a neural network such that the loss function is precisely this invariant.   This would be very effective because the minimum of the loss function would determine precisely the value of the zero map and therefore the transformation needed to get the Julia reduction of the form.  There is one major problem with this approach.  As Beshaj showed in her thesis
 \cite{beshaj-thesis} computation of the  Julia invariant  symbolically is extremely difficult even for 
 small degree forms such as quartics, quintics, and sextics.  
 
 However, geometrically this can be done rather easily for each $n$-gon as we illustrate next.  We can numerically compute the roots of $f(x)$ in the hyperbolic plave $\mathbb{H}_2$ including the real roots.  Using the method described in \cref{hyb-sec} we can find the hyperbolic centroid of such roots.  Even though this is computed numerically, we can always estimate a matrix $M \in \Sl_2 (\Z)$ such that the hyperbolic centroid is in the fundametal region $\mathcal{F}$. 

\subsection{Triangles and binary sextics}
We constructed a database of triangles for $r_1=1$ and $r_2=20$. There are  37 090 735 such triangles in $\mathbb{H}_2$.  The data is organized in a dictionary 
as:
\[
(\a_1, \a_2, \a_3):  [  [c_0, \ldots , c_6], [x_1, y_1], [ x_2, y_2]        ],
\]
where $(\a_1, \a_2, \a_3)$ is the key, $[c_0, \ldots , c_6]$ are the coefficients of $f(x, y)$,  $[x_1, y_1]$ is the center of mass, and 
$ [ x_2, y_2]  $ the hyperbolic centroid.

Among all such triangles  we are interested on the ones where the distance between the center of mass and the hyperbolic centroid are the biggest.  Out of $37\, 090\, 735$ such triangles, the one where this distance is maximum is for the triangle with vertices 
\[
\a_1=1+19 \iu , \a_2 = 2+19\iu, \a_3 = 19+ \iu.
\]
 The corresponding sextic is 
\begin{small}
\[
\begin{split} 
& f(x, y)  =   (x^2-38x+362)(x^2-4x+365)(x^2-2x+362)   =    x^6 -44x^5y  \\
& + 1325x^4 y^2  -32280x^3 y^3  + 480964x^2 y^4    -5809376x y^5  + 47831060 y^6
\end{split} 
\]
\end{small}
with height $\h (f) = 47 \, 831 \, 060$.   

\begin{figure}[h!] 
   \centering
   \includegraphics[width=2.5in]{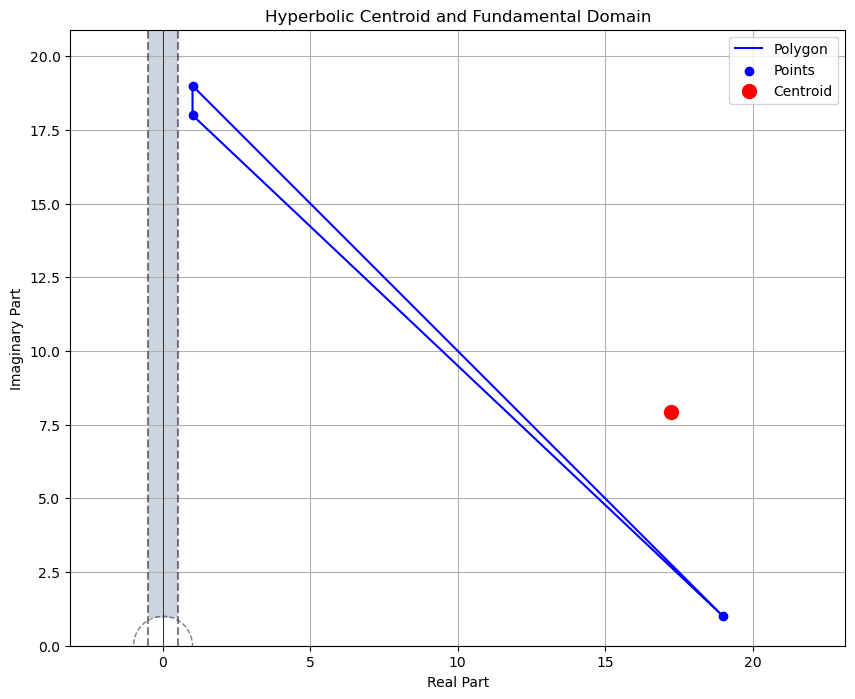} 
   \caption{The triangle with the maximum distance between the center of mass and hyperbolic centroid among 
     $37\, 090\, 735$ triangles.}
   \label{hyb-3}
\end{figure}

The center of mass  has coordinates  $ \left( \frac {22} 3, 13\right)$ and the hyperbolic center  $\left( \frac {52} 3, \frac {271} {100}  \right)$. 
To shift the center of mass to the fundamental domain we shift by seven units to the left ($z\to z-7$) which correspond to the matrix $M= \begin{bmatrix} 1 & - 7 \\ 0 & 1\end{bmatrix}$.  The Julia reduced form of  $f(x, y)$ is $f^{M^{-1}}= f(x+7, y)$, which is 
%
\[
\begin{split} 
\Jul{f}  & =  (x^2-24x+145) (x^2+10x +386)(x^2+12x +397)  
\end{split} 
\]
%
with height $\h (\Jul{f} ) = 22 \, 220 \, 090$, a significant improvement from the original height.    The hyperbolic reduction would correspond to the shifting $f(x+17, y)$ and give 
\[
\begin{split} 
\hyp{f}  & =  (x^2-4x+5) (x^2+30x +586)(x^2+32x +617)  
\end{split} 
\]
with a height $\h (\hyp{f}) = 1 \, 807 \, 810$, a significant improvement from the Julia reduction.  This is the first example that we know where the hyperbolic reduction gives a much smaller height than the Julia reduction. 

However, something amazing happens here.  The height continues to get smaller if we shift to the left and achieve it minimum for $f(x+19, y)$, where the form becomes
 \begin{small}
\[
\begin{split} 
f(x+19, y)   & =  (x^2+1) (x^2+34x + 650) (x^2+36 x + 685)   
\end{split} 
\]
\end{small}
which has height $\h (f(x+19), y)  = 447\,  809$.    Is this the minimal height in the $\Gamma$-orbit?  Or we could ask even more, is this the minimal absolute height (i.e. the smallest height among all $\Gamma$-orbits)?    Notice that no transformation via diagonal matrices would lower the height here; see \cite{2020-1}.  Hence, very likely this is the minimal height.

A similar example where the Julia reduction was computed algebraically was given in \cite[Example 1]{min-beshaj}.  For the same example in \cite{e-sh} the hyperbolic center was computed and shown that was different from the center of mass.  However, they were too close to each other that the reduction both ways held the same result.  That was the reason that we looked though our large database for the example were the distance between the two centroids was maximum.  

In \cite{min-beshaj} was also shown that for binary sextics with extra involutions the center of mass was always in the $\iu$-axis. That is because such sextics have roots symmetric to this axis.  To avoid such cases of reducible forms we picked our triangles to by always with positive real part.  

As far as we are aware, this is the first  example where the two reductions are shown to give different results.  
This example shows that neither Julia reduction, nor hyperbolic reduction achieve minimal height.  Moreover, it seems to suggest that the hyperbolic reduction is a more natural approach since it preserves better the geometry of the hyperbolic plane.   

We compared both reductions from  all the data for triangles between circles $r=1$ and $r=5$ and found out that from all 518 665 binary sextics we have:
\[
\begin{aligned}
&\text{Hyperbolic reduction:}  & 270\, 997 \\
&\text{Julia reduction:} & 75\, 993   \\
&\text{Same result:}  & 171\, 675 \\
\end{aligned}
\]
Hence,  in this case the hyperbolic reduction performs considerably better than the Julia reduction.    This suggests that some mixture of the two methods might be more suitable.  Next, we will see how each reduction performs 
in the case of binary decimics.

\subsection{Pentagons  and binary decimics.}
We follow the same approach with the same assumptions as for the case of triangles.  Hence, we want to build a database of pentagons with vertices in the hyperbolic plane and with $\re (z) >0$.  
Since there will be more possible combinations in this case, we only for radius up to $r\leq 7$.  For each one of such pentagons we have a totally complex binary decimic.

\begin{figure}[h!] 
   \centering
   \includegraphics[width=3in]{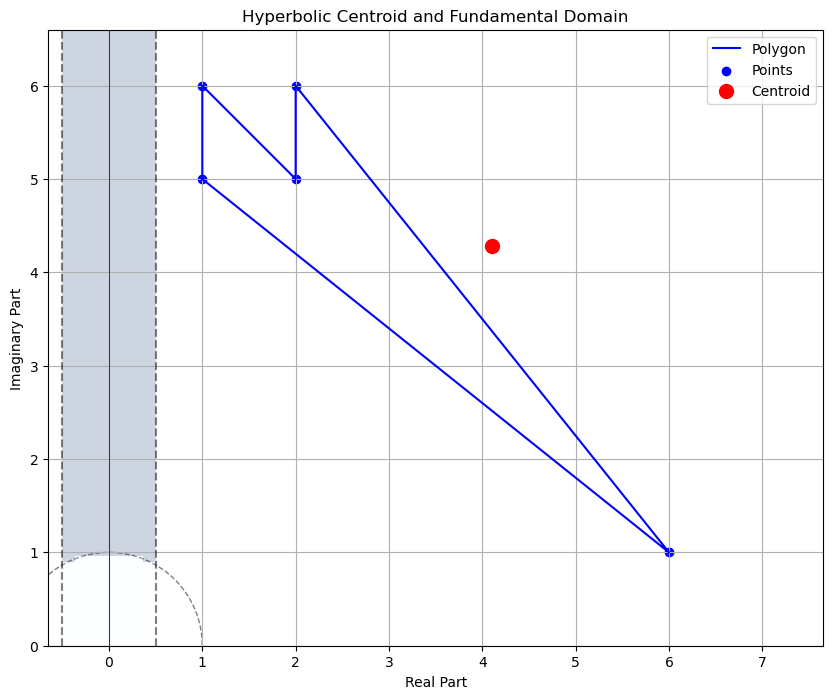} 
   \caption{The pentagon with the maximum distance between the center of mass and hyperbolic centroid among  $8 \, 936 \, 928$   pentagons.}
   \label{fig6}
\end{figure}

 In \cref{fig6} we graph the pentagon where such distance is the maximum between all $8 \, 936 \, 928$   pentagons for $r_1=1$ and $r_2=7$. 
It belongs to roots 
\[
1+5 \iu, \;   1+6 \iu, \;    2+6 \iu, \;   3+3 \iu, \;   6+\iu.
\]
 The corresponding degree ten binary form has height 
\[ \h (f) = 25 \, 627 \, 680 \]
The center of mass has coordinates $(2.6, 4.8)$ and the hyperbolic centroid is $(4, 24, 2. 94)$. The shift corresponding to the Julia reduction  (resp. hyperbolic reduction)  is $f(x+3, y)$ 
(resp. $f(x+4, y)$ )  
and has height   $\h (  Jul(f) )   =  3 \, 862\, 800$     (resp.  $\h (\hyp{f} ) = 3 \, 060\, 000$). 
Hence, again the hyperbolic reduction gets a better height, but not the minimal height, which is obtained for $f(x+5, y) $ and it is $h = 2 \, 494, 440$. The minimal polynomial is 
\[
f (x)= (x^2-2x+2) (x^2+4x+13) (x^2+6x+45)(x^2+8x+52)
\]
Hence, it seems as what is happening is very similar  as in the case of the triangles in the sense that it is closer to the far right vertex of the $n$-gon, but it does not exactly at the vertex as in the case of triangles.

We compared both reductions from  all the data for triangles between circles $r=1$ and $r=4$ and found out that from all  11 628 binary decimics we have:
\[
\begin{aligned}
&\text{Hyperbolic reduction:}  & 2\, 367 \\
&\text{Julia reduction:} & 797   \\
&\text{Same result:}  & 8\, 464 \\
\end{aligned}
\]
Hence,  again the hyperbolic reduction performs considerably better than the Julia reduction.      For the strip $r_1=1$ and $r_2=5$   we get 278 256 sextics from which 
\[
\begin{aligned}
&\text{Hyperbolic reduction:}  &81   \, 034 \\
&\text{Julia reduction:} &  33 \, 213  \\
&\text{Same result:}  &  164 \, 009 \\
\end{aligned}
\]

\def\p{\mathfrak p}
\def\wgcd{\mbox{wgcd} }
\subsection{Finding the minimal form}  
From the above work is clear that none of the two methods will determine the minimal form in every case.  Moreover, even one method is better than the other, it does not mean that is has reached the minimal form.  
There are two types of transformations that could be used to decrease the height of binary forms: \emph{shifts} and \emph{rescaling}.  Shifts,   $f(x+d)$ which have been discussed above will send a monic polynomial to a monic polynomial, therefore $f(x+d)$ is also primitive.  However, transformations $x \to \lambda x$ for the appropriate nonzero $\lambda$ can produce a binary form $f(\lambda x)$ which is not primitive.  This new binary form can have smaller height less that the original forms, since when we compute the height we must divide by the content of $f(\lambda x)$. \\

\noindent \textbf{Shifts:}  After shifting the form using the "best" reduction from Julia reduction or hyperbolic reduction above, we might need another \emph{additional shift} to reach the minimal height. For our experiments above for an additional shift $d\leq 3$ we always reach the minimal form, but most likely this is due to the size of our data.   It is an open problem to bound the size of this additional shift.  In the next section we will design a layer which determines this additional shift.  It is based on the fact that while $d$ is increasing in absolute value  then the height of $f(d \pm i)$ decreases until it reaches the minimum value and then it starts increasing again. \\

\noindent \textbf{Scaling: }   We can lower the height of the binary form by transformations of the form $x\to \lambda x$.  This used the fact that the height of the binary form $f=[a_0: \ldots : a_n]$ is the maximum of the absolute values $|a_i|$ when the form is primitive.  Hence, if we pick $\lambda$ such that  we maximize the greatest common divisor of the coefficients.  This was considered in \cite{min-beshaj} and also in \cite{2020-1} in terms of diagonal matrices.   We will take a slightly different approach here. 

Let  $f=[a_0: a_1 : \ldots : a_n]$.  Let $\w=(0, 1, 2, \ldots, n)$ be a set of heights.  Consider
$
\p_f= [a_1: \ldots : a_n]  \in \mathbb{P}_\w
$
 as a point in the weighted projective space  $\mathbb{P}_\w$.  We will denote by $\wh_\w (\p_f)$ the weighted height of $\p_f$   and by $\wgcd_\w (a_1, \ldots , a_n)$ the weighted greatest common divisor with respect to the weights $\w$;    see  \cite{2019-1}  for details.  
%
 Let $m = \mbox{lcm} (1, \ldots , n)$. Then,  it was proved in \cite{2019-1} that 
\[  \wh (a_1, \ldots , a_n) < \h \left( (a_1, \ldots , a_n) \right)^{\frac 1 m}
\]
%
%
\begin{lem}\label{scaling}
Let $f$ and $w$ as above
and $p$ and $q$ integers such that 
\[
p= \gcd (a_0, q). \quad \text{ where } q = \wgcd_\w   (a_1, \ldots , a_n)
\]
Then  $f \left( p x, y \right)$ has the minimum height that can be achieved by scaling
\end{lem}

\proof
Suppose that there is $g(x, y)$ which is obtained from  $f(x, y)$ by scaling and  has smaller height.  Then, there exists a non-zero $\lambda$ such that 
$f( \lambda x, y) = g(x, y)$. Hence,   coefficients change as
\[
f= [a_0: \ldots : 1] \to [a_0: \lambda a_1 : \ldots : \lambda^{n-1} a_{n-1} : \lambda^n] =g
\]
That means that $\lambda | q$ and since we are assuming that the height of $g$ is smaller than $f$, that implies that $\lambda | a_0$.  Hence, $\lambda | p$. That completes the proof. 
\qed

Lemma above provides an algorithm how to get the form with minimum height and will b implemented as the \emph{scaling layer} in the next section.

\section{A Machine Learning Approach to Reducing Binary Forms}  
Determining the transformation that reduces a binary form to its minimal height remains an open and challenging problem. 
Historically, Julia reduction was considered the most effective method of reduction for binary forms. It generalized the reduction of quadratics, which successfully minimizes the discriminant and the height. This motivated attempts to generalize  reduction to higher-degree forms. However, in contrast to quadratics, higher-degree forms involve multiple invariants, making the minimization problem more complex. Minimizing these invariants, referred to as \emph{Reduction A} in \cite{2020-1}, can be achieved using weighted greatest common divisors and weighted heights, as discussed in \cite{2022-1, 2023-01}.   However, minimizing the invariants does not necessary means minimizing the coefficients, which is a  complex arithmetic problem.

Despite the progress made by these approaches, neither Julia reduction nor hyperbolic reduction guarantees achieving the minimal form for binary forms of arbitrary degree. To address this limitation, we propose a novel machine learning framework designed to predict transformations that effectively minimize the height of binary forms. Our approach combines neural networks with symbolic layers to improve the model's accuracy and interpretability.  

\subsection{Architecture of the Model}  

The input to the model is a degree \(n \geq 2\) binary form, represented as a projective point:  
\[
[ c_0 : \ldots : c_n ] \in \mathbb{P}^n.
\]
The model is composed of the following layers:  \\

\noindent\textbf{Roots layer:}  
In this layer, we compute the roots of the binary form in the upper half-plane \(\mathbb{H}_2\) numerically. This provides the essential geometric data for subsequent computations. 
The Python code for this computation is provided  below.\\

\begin{lstlisting}[language=Python]
def roots_upper_half_plane(f, precision=10):
    f_poly = sp.Poly(f, x)
    roots = f_poly.all_roots()
    upper_half_roots = []
    for r in roots:
        r_num = sp.N(r, precision)
        if hasattr(r_num, 'imag') and r_num.imag > 0:   
            upper_half_roots.append(r_num)
    return upper_half_roots
\end{lstlisting}

\bigskip

\noindent\textbf{Hyperbolic layer:}  
This layer computes the hyperbolic centroid of the roots in \(\mathbb{H}_2\) using the formula from \cref{hyb_centroid}. The centroid serves as a geometric invariant that guides the reduction process. \\

 \begin{lstlisting}
def hyperbolic_centroid(points):
    x_coords = [x for x, _ in points]
    y_coords = [y for _, y in points]
    n = len(points)
    s_n_minus_1 = sum(prod(y_coords[:i] + y_coords[i + 1:]) for i in range(n))
    t = sum(prod(y_coords[:i] + y_coords[i + 1:]) * x_coords[i] for i in range(n)) / s_n_minus_1
    norm_squared = sum(
        prod(y_coords[:i] + y_coords[i + 1:]) * (x_coords[i] ** 2 + y_coords[i] ** 2)
        for i in range(n)
    ) / s_n_minus_1
    u = (norm_squared - t**2) ** 0.5
    return round(t, 2), round(u, 2)
\end{lstlisting}

\bigskip

\noindent\textbf{Direction layer:}  
While the hyperbolic centroid provides a useful geometric indicator, it does not guarantee the minimal form. The direction layer determines the optimal shift direction in \(\mathbb{H}_2\) to further reduce the height of the binary form. This step refines the reduction process by identifying the transformation that leads to the most significant height reduction.  \\

\begin{lstlisting}
def direction_layer(L):
    for key, value in L.items():
        coefficients = [int(c) for c in key.strip('[]').split(',')]
        original_height = height(coefficients)
        transformed_plus = [
            sum(coefficients[j] * comb(j, i) for j in range(i, len(coefficients)))
            for i in range(len(coefficients))
        ]
        height_plus = height(transformed_plus)
        transformed_minus = [
            sum(coefficients[j] * (-1)**(j - i) * comb(j, i) for j in range(i, len(coefficients)))
            for i in range(len(coefficients))
        ]
        height_minus = height(transformed_minus)
        if height_plus < original_height:
            value.append('+')
        if height_minus < original_height:
            value.append('-')
    return L
\end{lstlisting}

\bigskip

\noindent\textbf{Scaling layer:}  
The scaling layer handles reductions up to \(\mathrm{GL}_2(\mathbb{Q})\)-equivalence by applying a scaling transformation of the form \(x \to \lambda x\) for some \(\lambda \in \mathbb{Q}^\star\). This step ensures that the resulting binary form achieves a minimal height with respect to \(\mathrm{GL}_2(\mathbb{Q})\)-equivalence. The theoretical basis for this layer is provided by the scaling lemma (see \cref{scaling}).  \\

\begin{lstlisting}
def scaling_layer(key):
    from math import gcd
    def divisors(x):
        x = abs(x)
        return [i for i in range(1, x + 1) if x % i == 0]
    key = primitive(key)
    coefficients = key   
    n = len(coefficients) - 1   
    c_0 = coefficients[0]
    possible_d = divisors(c_0)   
    for d in possible_d[1:]:  
        if c_0 % (d ** n) != 0:
            continue   
        valid = True
        for i, c_i in enumerate(coefficients):
            if c_i % (d ** (n - i)) != 0:
                valid = False
                break
        if valid:
            g = [c * (d ** i) for i, c in enumerate(coefficients)]
            return primitive(g)
    return key
    
\end{lstlisting}

\medskip

Having introduced the layers of the machine learning model, we now turn to the details of its implementation and the challenges encountered during training.

\subsection{Implementation Details}  

Our implementation is designed to handle binary forms of various degrees, including degrees 5, 6, and 10, with detailed databases described in \cref{data}. All datasets and code will be made publicly available. 

Initial attempts to use unsupervised machine learning models achieved low accuracy rates of 10--20\%. However, the inclusion of symbolic layers significantly improved performance, demonstrating the value of combining neural networks with symbolic computation. 

A major challenge in training the model was the lack of reliable, large-scale datasets for higher-degree binary forms that include their corresponding minimal forms. While it is straightforward to generate large datasets of binary forms, these datasets often lack the necessary ground truth for minimal reductions. To overcome this, we employed alternative methods to construct training data, combining algorithmic reduction techniques with symbolic computations to approximate minimal forms.


\section{Conclusions}

Binary forms have been the focus of classical mathematics and continue to be the focal point of current research; see \cite{2024-07, 2020-1, 2024-02, 2019-1, 2022-1, bhargava-2015}. 
This study provides a comparative analysis of Julia reduction and hyperbolic reduction for finding equivalent binary forms with minimal coefficients. Our results demonstrate that hyperbolic reduction generally achieves better outcomes than Julia reduction, particularly for sextics and decimics. However, neither method guarantees achieving the minimal form, highlighting the need for additional transformations. To address this, we introduced an additional shift and scaling approach that further reduces the form, offering an improved but not absolute solution.

A significant contribution of this work is the proposal of a machine learning framework to determine optimal transformations. This approach bridges traditional mathematical methods with data-driven techniques, offering a novel perspective on the problem. The success of this framework suggests that machine learning can be a valuable tool in exploring the complex landscape of binary forms, particularly in identifying patterns and relationships that are difficult to capture through classical methods alone.

Despite these advancements, certain limitations remain. Both Julia and hyperbolic reduction methods are heuristic in nature and do not guarantee a minimal form, and the proposed machine learning framework requires further development to generalize across a wider range of forms. Additionally, the reliance on computational experiments necessitates high computational resources, which may limit the scalability of the methods.

Looking forward, there are several promising directions for future research. First, enhancing the machine learning framework with larger and more diverse training datasets could improve its robustness and accuracy. Second, exploring connections between reduction theory and other areas of computational mathematics, such as lattice reduction or invariant theory, may yield new insights. Finally, developing theoretical guarantees for achieving minimal forms under specific conditions remains an open and intriguing question.

This work lays a foundation for integrating classical reduction techniques with modern computational tools, offering both practical solutions and a deeper understanding of binary forms. By combining traditional methods with machine learning, we take a step toward more effective and generalizable approaches to symbolic computation and reduction.

\bibliographystyle{unsrt}
\bibliography{julia}

\end{document}